\documentclass[sigconf]{acmart}
\AtBeginDocument{%
  \providecommand\BibTeX{{%
    \normalfont B\kern-0.5em{\scshape i\kern-0.25em b}\kern-0.8em\TeX}}}

\usepackage{float}
\usepackage{stfloats}
\usepackage{booktabs, siunitx}
\usepackage{multirow}
\usepackage{xspace}
\usepackage{subcaption}
\usepackage{pgfplots}
\pgfplotsset{compat=newest}
\usepackage{makecell}
\usepackage{rotating}
\usepackage{afterpage}
\usepackage{titlesec}

\titleformat{\subsection}
  {\normalfont\large\bfseries}{\thesubsection}{1em}{}

\copyrightyear{2024}
\acmYear{2024}
\setcopyright{acmlicensed}\acmConference[HRI '24]{Proceedings of the 2024 ACM/IEEE International Conference on Human-Robot Interaction}{March 11--14, 2024}{Boulder, CO, USA}
\acmBooktitle{Proceedings of the 2024 ACM/IEEE International Conference on Human-Robot Interaction (HRI '24), March 11--14, 2024, Boulder, CO, USA}
\acmDOI{10.1145/3610977.3634975}
\acmISBN{979-8-4007-0322-5/24/03}
    
\begin{document}

\title{Feel the Bite: Robot-Assisted Inside-Mouth Bite Transfer using Robust Mouth Perception and Physical Interaction-Aware Control}

\author{Rajat Kumar Jenamani}
\affiliation{%
  \institution{Cornell University}
}
\email{rj277@cornell.edu}

\author{Daniel Stabile}
\affiliation{%
  \institution{Cornell University}
}

\author{Ziang Liu}
\affiliation{%
  \institution{Cornell University}
}

\author{Abrar Anwar}
\affiliation{%
  \institution{University of Southern California}
}

\author{Katherine Dimitropoulou}
\affiliation{%
  \institution{Columbia University}
}

\author{Tapomayukh Bhattacharjee}
\affiliation{%
  \institution{Cornell University}
}

\renewcommand{\shortauthors}{Rajat Kumar Jenamani et al.}


\def\note#1{\textcolor{blue}{#1}} 
\def\tnote#1{\textcolor{red}{Tapo: #1}} 
\def\rkjnote#1{\textcolor{cyan}{Rajat: #1}} 
\def\dsnote#1{\textcolor{orange}{Daniel: #1}} 

\def\kathleen#1{P1}
\def\tucker#1{P2}
\def\aimee#1{P3}
\def\ron#1{P4}
\def\benjamin#1{P5}
\def\paul#1{P6}
\def\laurel#1{P7}
\def\luke#1{P8}
\def\julia#1{P9}
\def\atoniotte#1{P10}
\def\nyree#1{P11}
\def\yessenia#1{P12}
\def\tabhita#1{P13\xspace}

\begin{abstract}
\vspace{-0.1cm}

Robot-assisted feeding can greatly enhance the lives of those with mobility limitations. Modern feeding systems can pick up and position food in front of a care recipient's mouth for a bite. However, many with severe mobility constraints cannot lean forward and need direct inside-mouth food placement. This demands precision, especially for those with restricted mouth openings, and appropriately reacting to various physical interactions — incidental contacts as the utensil moves inside, impulsive contacts due to sudden muscle spasms, deliberate tongue maneuvers by the person being fed to guide the utensil, and intentional bites. In this paper, we propose an inside-mouth bite transfer system that addresses these challenges with two key components: a multi-view mouth perception pipeline robust to tool occlusion, and a control mechanism that employs multimodal time-series classification to discern and react to different physical interactions. We demonstrate the efficacy of these individual components through two ablation studies. In a full system evaluation, our system successfully fed 13 care recipients with diverse mobility challenges. Participants consistently emphasized the comfort and safety of our inside-mouth bite transfer system, and gave it high technology acceptance ratings — underscoring its transformative potential in real-world scenarios. Supplementary materials and videos can be found at: \href{http://emprise.cs.cornell.edu/bitetransfer/}{\textbf{emprise.cs.cornell.edu/bitetransfer}}.

\vspace{-0.15cm}

\end{abstract}

\begin{teaserfigure}
 \centering
  \vspace{-0.4cm}
  \includegraphics[width=\textwidth]{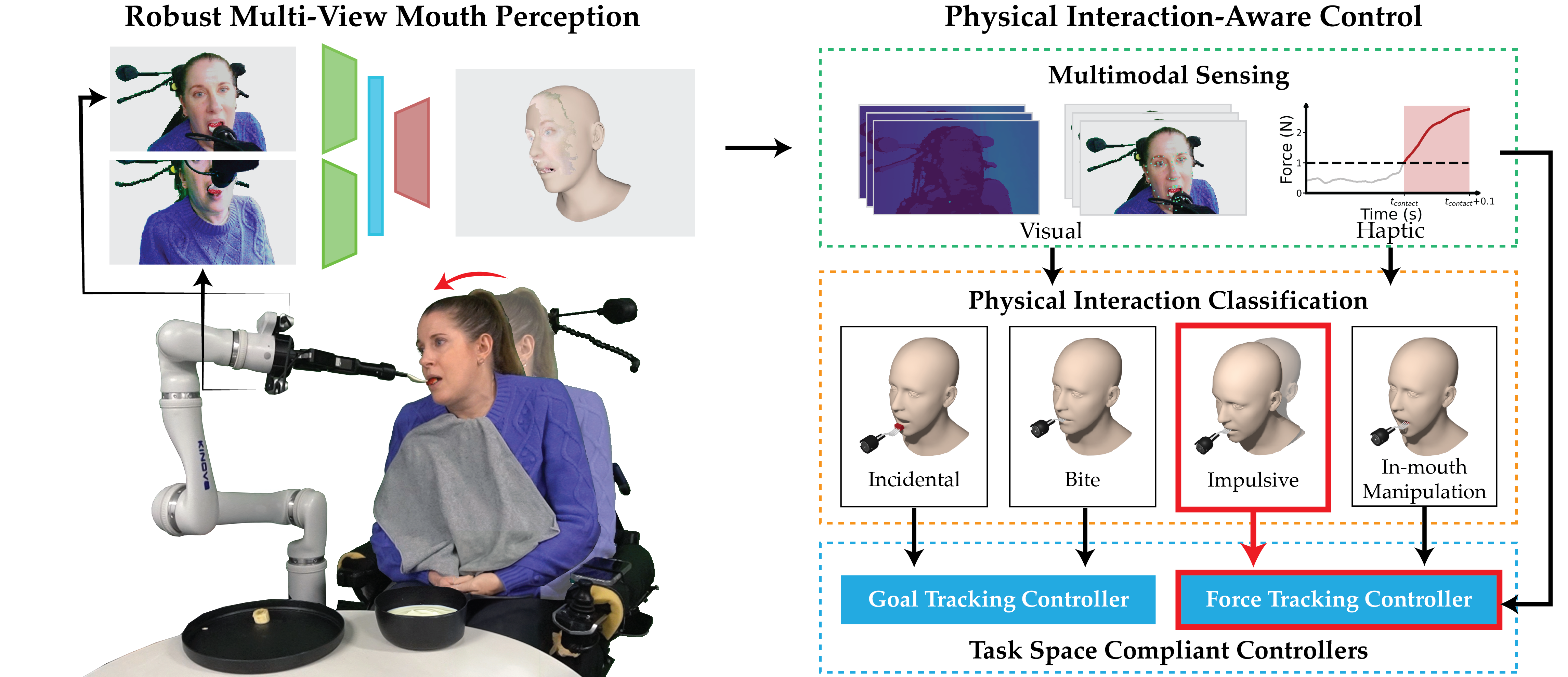}
  \vspace{-0.7cm}
  \caption{We propose an inside-mouth bite transfer system that uses two key components, (i) robust multi-view mouth perception, and (ii) physical interaction-aware control, to successfully feed care recipients with diverse mobility limitations.}
  \vspace{-0.15cm}
  \label{fig:teaser}
\end{teaserfigure}

\ccsdesc[500]{Human-centered computing~Accessibility technologies}
\ccsdesc[500]{Computer systems organization~Robotics\vspace{-0.25cm}}

\keywords{Assistive Robots, Physical Human-Robot Interaction, Robot-Assisted Feeding
}


\maketitle

\section{INTRODUCTION}

Eating is an Activity of Daily Living (ADL) \cite{katz1963studies}. Losing the ability to independently feed oneself can be devastating, and may elicit feelings of shame and dependence among care recipients \cite{jacobsson2000eatingprocess, shune2020experience, nanavati2023design}. In the United States alone, approximately one million adults with mobility limitations need assistance with eating \cite{taylor2018americans}. Feeding is also one of the most time-consuming ADLs for caregivers \cite{chio2006caregiver}, placing a significant burden on them \cite{dreer2007family, lynch2017impact}. Robot-assisted feeding systems have the potential to enhance the quality of life of care recipients \cite{brose2010role} and reduce caregiver burden. 

Robot-assisted feeding can be broken down into two stages: bite acquisition and bite transfer \cite{madan2022sparcs}. Bite acquisition involves acquiring a bite-sized food item with an appropriate utensil \cite{feng2019robot, gordon2020adaptive, gordon2021leveraging, gordon2023towards}, whereas bite transfer entails moving it from above the plate to the care recipient's mouth \cite{gallenberger2019transfer, belkhale2021balancing, shaikewitz2023mouth}. In this paper, we focus on bite transfer. Most prior works bring
a food item in front of the mouth of a care recipient, who is then expected to lean forward to take the bite. However, for those with severe upper body and neck mobility limitations, leaning forward for an outside-mouth bite transfer can be functionally impossible, necessitating the direct placement of food in their mouths. Even for care recipients who can lean forward, having to repeatedly make this movement can be exhausting. 
In this work, we present a robot-assisted feeding system that can perform inside-mouth bite transfer and demonstrate its utility for assisting a diverse group of care recipients with severe mobility limitations.

Care recipients who need inside-mouth bite transfers often have complex medical conditions -- including limited mouth opening \cite{GRANGER1999697}, involuntary movements (spasms), and requirement of food transfer at a specific location inside their mouth -- which makes feeding them extremely challenging. Involuntary motions, diverse in their type and occurrence, demand alert and adaptable feeding strategies. For example, if an involuntary forward spasm occurs just as the utensil approaches the care recipient's mouth, the caregiver must quickly retract the utensil to avoid harm. If such a motion happens with the utensil already inside the mouth, the caregiver should comply with the motion, ensuring the utensil moves in harmony with it to avoid injury. Another layer of intricacy emerges when food must be placed precisely within the mouth at a preferred transfer location. In these instances, care recipients may use their tongue to guide the utensil, necessitating that caregivers recognize and act upon this cue. In addition to this, physical interactions also occur when slight errors in sensing/control lead to incidental contacts, and when care recipients intentionally bite down on the food. Consequently, feeding individuals with such complex needs requires precision in perception and control, and identifying and appropriately reacting to various types of physical interactions. 

In this paper, we make the following contributions: 

\textbf{Inside-Mouth Bite Transfer System.} We propose a system that leverages two key components -- robust multi-view mouth perception, and physical interaction-aware control -- to address the aforementioned challenges and successfully feed care recipients with severe mobility limitations food inside their mouth.

\textbf{Robust Multi-View Mouth Perception Method.} 
Existing methods for mouth perception in robot-assisted feeding rely on a single in-hand camera view, and thus face challenges during inside-mouth bite transfer due to significant occlusion from the feeding utensil. To address this, we introduce a novel multi-view mouth-perception approach that is robust to tool occlusion (Section \ref{mouth_perception}). This method allows for uninterrupted real-time mouth perception throughout the transfer process, enabling detection and adaptation to both voluntary and involuntary head movements.



\textbf{Physical Interaction-Aware Control Method.} We propose an interaction-aware controller that uses multimodal sensing for classifying the nature of physical interactions in real-time and reacts accordingly (Section \ref{interaction_classification}). 
Adopting a data-driven approach, we collect a multimodal dataset comprising various types of physical interactions that can occur and train time-series classification models. 
Based on the detected interaction type, we switch between a goal-tracking controller and a force-tracking controller.  

\textbf{Evaluation of System Components.} We demonstrate the necessity of both novel components through evaluations with baselines and two ablation studies involving participants without mobility limitations (Section \ref{study_mouth_perception}).

\textbf{Full System Evaluation with Care Recipients.} We demonstrate our system's efficacy through a user study with 13 care recipients with diverse medical conditions, all necessitating assistance in feeding. Findings suggest users perceive our system as safe and comfortable, and view the technology favorably as measured using a Technology Acceptance Model (TAM) survey \cite{davis1989user}. 
\vspace{-0.1cm}
\section{RELATED WORK}

\hspace{\parindent} \textbf{Robot-Assisted Feeding.} While many commercial feeding systems \cite{obi, neater} exist, their limited autonomy and need for manual trajectory programming for bite transfer hinder widespread adoption and retention. Over recent years, various robot-assisted feeding systems with autonomous transfer capabilities have been proposed \cite{gallenberger2019transfer, candeias2018vision, belkhale2021balancing}. However, they assume that care recipients can lean forward, causing the robot to stop at a predetermined distance from their mouth. 
More recently, a few systems have explored inside-mouth bite transfer \cite{park2020evaluation, shaikewitz2023mouth}, but they perceive the user's mouth pose only once (initially) and do not continuously track it during transfer. 
This requires users to remain static throughout the entire transfer process, a challenging demand for many individuals
with pronounced mobility impairments. These systems also do not consider various types of physical interactions that can arise during
inside-mouth bite transfer, such as incidental contacts, impulsive contacts, and in-mouth manipulation. To the best of our knowledge,
our work is the first to demonstrate autonomous inside-mouth bite transfer for care recipients having complex medical conditions such
as limited mouth opening, involuntary movements, and precise food placement requirements for feeding. 

\begin{figure*}[tp]
  \includegraphics[width=\textwidth]{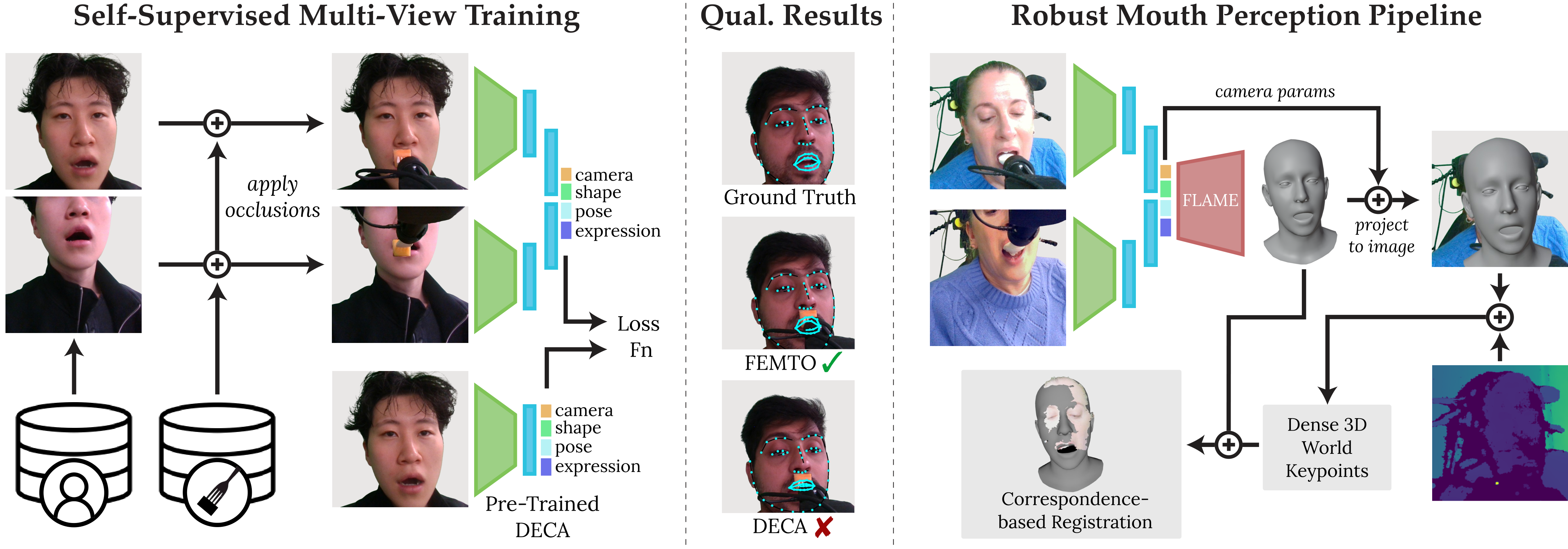}
  \vspace{-0.8cm}
  \caption{The self-supervised multi-view training of FEMTO (left) enables robust mouth perception under tool occlusion (center). Our perception pipeline (right) leverages FEMTO to accurately estimate mouth pose despite noisy depth due to occlusion.
  }
  \vspace{-0.6cm}
  \label{fig:head_perception}
\end{figure*}

\textbf{Mouth Perception.} 
To effectively feed care recipients with complex medical conditions, it is necessary to accurately perceive their mouth for the whole duration of inside-mouth bite transfer.
Contemporary robot-assisted feeding systems use single in-hand cameras, and localize the mouth by either projecting 2D mouth landmarks onto an aligned depth image \cite{shaikewitz2023mouth}, or fitting a fixed-head model to sparse 3D facial landmarks \cite{gallenberger2019transfer}. However, near the mouth, significant occlusion from the utensil leads to noisy depth data and inaccuracies in single-view perception of 2D landmarks, rendering these methods ineffective for continuous mouth tracking. In this work, we propose a mouth perception pipeline that leverages multiple in-hand cameras and parameterized head models to be robust to tool occlusion, thus enabling uninterrupted and accurate mouth perception. Among parameterized head models \cite{booth20163d, dai20173d, li2017flame}, Faces Learned with an Articulated Model and Expressions (FLAME) \cite{li2017flame} is of special interest as it separates the representation of identity, pose, and facial expression. Detailed Expression Capture and Animation (DECA) \cite{feng2021learning}, is state-of-the-art for FLAME parameter estimation, and takes a single RGB image as input. However, this reliance on monocular RGB images makes DECA vulnerable to inaccuracies in our use case where significant occlusions around the mouth are present. To overcome this limitation, we propose a novel encoder based on DECA's architecture that integrates data from multiple cameras, ensuring robust prediction of FLAME parameters.

\textbf{Physical Interaction-Aware Control.} Prior work in contact classification distinguishes between incidental collisions and intentional task contacts during collaborative human-robot manipulation 
\cite{kouris2018frequency,golz2015using,cioffi2020data,franzel2021detection,lippi2021data}. 
Classical approaches \cite{kouris2018frequency} use the spectral norm of external force/torque signals in a specific frequency range for classification. In contrast, several other methods leverage learning-based approaches like Support Vector Machines with time-series features derived from physical contact models \cite{golz2015using,cioffi2020data,franzel2021detection}, or time-series models such as RNNs \cite{lippi2021data}. While these methods typically use haptic data, we are inspired by works that integrate multiple sensing modalities \cite{lee2019making,fazeli2019see,li2019connecting,sundaresan2023learning}. To the best of our knowledge, our work is the first to explore multimodal (visual + haptic) sensing for physical human-robot interaction classification. It is also the first to categorize physical interaction types beyond incidental and intentional, by additionally identifying impulsive interactions and further distinguishing intentional interactions into in-mouth manipulation and bite interactions. 

\vspace{-0.1cm}
\section{Inside-Mouth Bite Transfer System}

Our inside-mouth bite transfer system (Figure \ref{fig:teaser}) consists of a Kinova Gen3 6 DoF robotic arm \cite{kinova} with a Robotiq 2F-85 gripper \cite{robotiq} grasping a custom-built feeding tool. It uses two Intel RealSense D415 RGBD cameras mounted on the robot’s wrist, one above and one below the utensil, for visual input, and a 6-axis ATI Nano25 F/T sensor \cite{ati} for haptic feedback. Our system leverages two key novel components - robust multi-view mouth perception (Section \ref{mouth_perception}) and physical interaction-aware control (Section \ref{interaction_classification}). 




\subsection{Robust Multi-View Mouth Perception}
\label{mouth_perception}
We require accurate mouth perception for the entire duration of inside-mouth bite transfer to successfully feed care recipients having small mouth openings, involuntary motions, and requirements of food placements at specific locations inside their mouth. However, real-time mouth perception using a single in-hand camera is challenging as there is significant occlusion from the utensil due to which: (i) state-of-the-art monocular methods such as DECA \cite{feng2021learning} fail at mouth keypoint detection, and (ii) even if we obtain the mouth keypoints, projecting them on the depth image for pose estimation in the real-world fails as the depth image is noisy in the vicinity of the utensil.
We propose a mouth perception pipeline that is robust to this occlusion challenge. Central to this pipeline is a novel method - Face Estimation from Multiple Views under Tool Occlusion (FEMTO). FEMTO uses inputs from two in-hand cameras mounted diametrically opposite to reconstruct a personalized head model for the care recipient, and estimates dense 2D facial keypoints. For the personalized head model, we use the parameterized head model FLAME \cite{li2017flame} which is represented by a small number of shape, pose, and expression parameters.

\textbf{Model Architecture.} FEMTO's architecture consists of two parallel encoders; one encoder for the top image and the other for the bottom. These encoders use frozen weights from DECA \cite{feng2021learning}, a single-view model pre-trained on 2 million images from large datasets \cite{cao2018vggface2, wang2019racial, chung2018voxceleb2}. The outputs of these encoders are concatenated and further processed through two fully-connected layers. These layers integrate information from both cameras, and, similar to DECA, generate FLAME parameters and an orthographic camera pose for projecting the 3D FLAME mesh into image space. 

\textbf{Self-Supervised Finetuning on Self-Curated Dataset.} To make FEMTO robust to occlusions, we finetune it on a self-curated visual dataset. We collect about 5000 unoccluded multi-view images with different head poses and facial expressions from 10 participants. Data from 8 participants are used for training FEMTO, and data from the remaining 2 participants are used for comparison against baselines. The latter is detailed in Section \ref{mouth_perception_evaluation}. These images are captured without the utensil in place, and thus DECA can typically generate accurate FLAME parameters and camera pose for these images. We then synthetically occlude these images by in-painting the utensil (Figure \ref{fig:head_perception}). We use various utensils holding distinct food items at this step to ensure that FEMTO can generalize to new utensils and food items. We provide these occluded images to FEMTO, and use annotations generated by DECA for the corresponding un-occluded image as ground truth for supervision. We evaluate all the generated ground truth parameters and manually update them wherever necessary. FLAME's modular parameterization allows for focusing training on parameters where DECA especially struggles under utensil occlusion, notably jaw pose prediction which is vital for detecting whether the mouth is open or closed.  

\textbf{Robust Mouth Perception Pipeline.} FEMTO processes the top and bottom images to generate FLAME parameters and the orthographic camera pose of the top RGB image relative to the head model (Figure \ref{fig:head_perception}). These FLAME parameters decode into a custom 3D head model, and the generated camera pose can be used to project this model to the top RGB image and obtain dense 2D facial keypoints. We combine these 2D keypoints with the aligned depth image from the top camera to get dense 3D world keypoints. Finally, we use a correspondence-based registration robust to outliers to pose the generated head model to the 3D world keypoints. Despite the depth data around the mouth region being noisy due to occlusion from the utensil, the customized head model enables us to use depth data from other regions of the head for accurate mouth pose estimation. This proposed pipeline runs real-time (5-10Hz) on a system with RTX 3090. 

Details on data collection, model training, and correspondence-based registration method used are in the Appendix \cite{appendix}. 

\vspace{-0.3cm}
\subsection{Physical Interaction-Aware Control}
\label{interaction_classification}
\vspace{-0.1cm}
Our physical interaction-aware control method first classifies the subtle physical interactions during bite transfer and switches appropriate compliant controllers accordingly. We take a data-driven approach to identify the nature of physical-interaction, and collect visual and haptic data for four types of physical interactions:
\begin{enumerate}
    \item Incidental interactions (collisions) between the utensil and user's mouth that occur outside the mouth due to sensing/control errors as the robot attempts to move inside.
    \item In-mouth manipulation interactions initiated by the user using their tongue to guide the utensil to a desired transfer location inside their mouth.
    \item Impulsive interactions due to involuntary spasms occurring while the utensil is inside the mouth.
    \item Bite interactions that occur when the user takes a bite.
\end{enumerate}

\textbf{Data Collection.} We collect a total of 3072 physical interactions (512 interactions X 6 participants without mobility limitations), varying both the utensil and the robot controller. For each physical interaction, we ask participants to perform different variations of the interaction. For example, when taking bites, participants are instructed to use only their teeth around the food item, only their lips around the utensil, or both. 


\textbf{Multimodal Classification.} For each physical interaction, we analyze visual and haptic data from a 100ms window starting at contact initiation. Visual features include 2D and 3D face keypoints generated from our mouth perception pipeline (Section \ref{mouth_perception}). In line with related haptic classification work \cite{golz2015using,franzel2021detection}, we augment raw haptic data with time-series features computed over the window, such as mean, range, kurtosis, Hjorth complexity, and frequency domain. We evaluate four models: Support Vector Machines (SVM) \cite{cortes1995support}, Multi-Layer Perceptron (MLP) \cite{haykin1994neural}, Temporal Convolutional Networks (TCN) \cite{lea2017temporal}, and Time-Series Transformers (TST) \cite{zerveas2021tst}. 



\textbf{Controller Switching.} Once the robot recognizes the nature of physical interaction, it switches to an appropriate controller using an event-driven control method. We use two compliant controllers during inside-mouth bite transfer: a goal-tracking controller and
a force-tracking controller. The goal-tracking controller tracks a given goal pose and is ideal for moving the food item to a desired
position inside the user’s mouth and moving the utensil outside their mouth as soon as a bite is detected. The force-tracking controller minimizes contact force on the F/T sensor at the end-effector and thus is very compliant and reactive. This is ideal for physical interactions such as impulsive and in-mouth manipulation, where the robot must quickly respond to forces applied to the utensil.


Details on data collection, model training, and controller implementation are provided in the Appendix \cite{appendix}.

\vspace{-0.2cm}
\section{Evaluation of System Components}
\vspace{-0.1cm}



\label{study_mouth_perception}

\subsection{Mouth Perception Evaluation}
\label{mouth_perception_evaluation}
\vspace{-0.1cm}

We compare the performance of proposed mouth perception method against baselines for accurate 3D mouth keypoint generation, and perform an ablation study to evaluate the necessity of a key feature our mouth perception pipeline enables: continuous, real-time mouth perception for inside-mouth bite transfer.

\noindent \textbf{Eval. 1: Comparison with Baselines -} We compare the root mean square error (RMSE) of 3D mouth keypoints generated by our novel mouth perception method (FEMTO) against a pre-trained DECA using our proposed pipeline, and contemporary mouth perception methods in robot-assisted feeding \cite{shaikewitz2023mouth, gallenberger2019transfer} (Table \ref{table_mouth_perception}). For this evaluation, we use data from the 2 test participants in Section \ref{mouth_perception}, introducing synthetic utensil occlusion to color and depth images. Ground truth keypoints are generated by applying our mouth perception pipeline with DECA to the unoccluded originals. 



\begin{table}[h]
\centering
\vspace{-0.3cm}
\caption{FEMTO outperforms baselines adapted from robot-assisted feeding \cite{shaikewitz2023mouth, gallenberger2019transfer} and head perception \cite{feng2021learning} literature.}
\vspace{-0.3cm}
\label{tab:robot_model_comparision}
\begin{tabular}{lr}
\toprule
\multicolumn{1}{c}{Method} & \multicolumn{1}{c}{RMSE (in $mm$)} \\ \hline
Gallenberger et al. 2019 \cite{gallenberger2019transfer} & $110.454$ \\
Shaikewitz et al. 2023 \cite{shaikewitz2023mouth} &  $82.946$ \\
DECA \cite{feng2021learning} &  $4.72$ \\
\textbf{FEMTO (Ours)} &  $\pmb{3.69}$ \\
\hline
\end{tabular}
\label{table_mouth_perception}
\vspace{-0.3cm} 
\end{table}


\begin{figure*}[t]
    \centering

    \includegraphics[width=\textwidth]{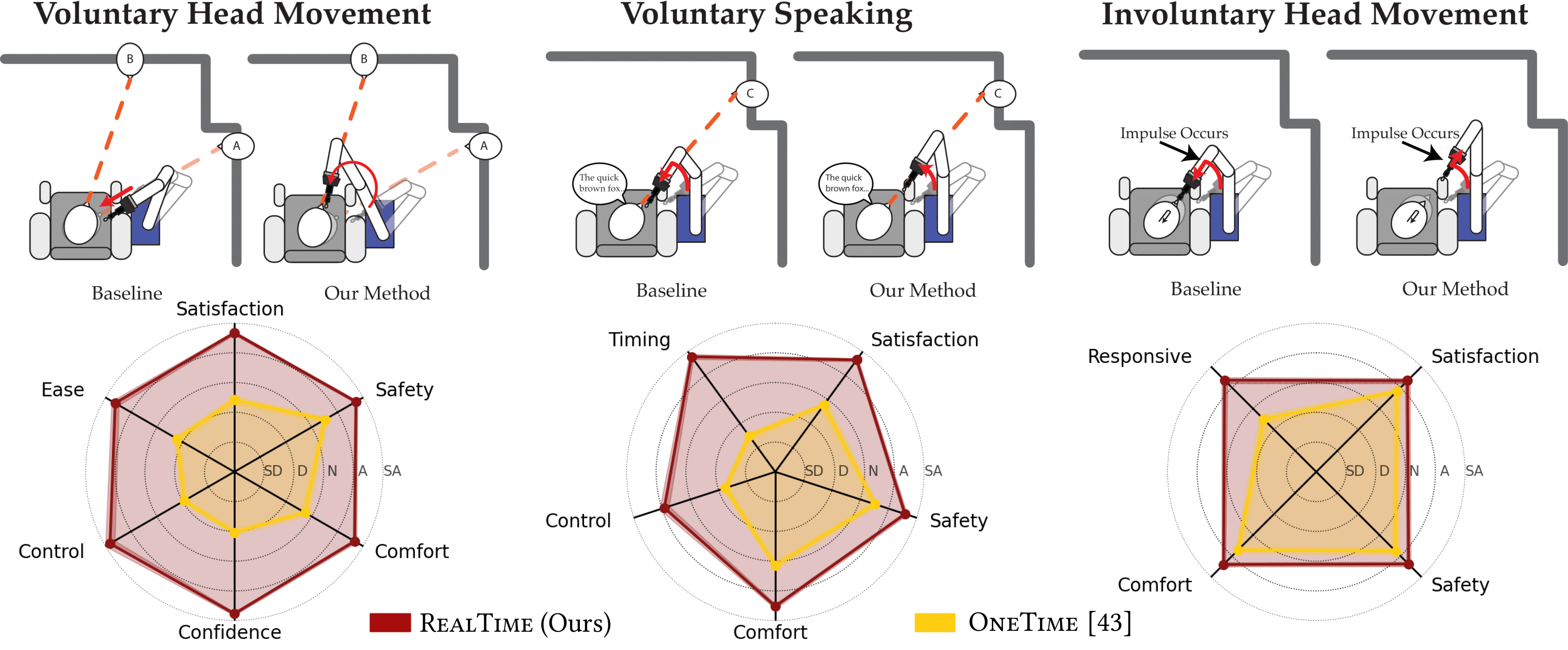}
    \vspace{-0.9cm}
    \caption{\textsc{RealTime} (Ours) enables the robot to track the mouth's pose and state, which enhances satisfaction, safety, comfort, confidence, control, ease of use, bite timing, and responsiveness across various scenarios in comparison to \textsc{OneTime} \cite{shaikewitz2023mouth}. Response options used across studies: SD (Strongly Disagree), D (Disagree), N (Neutral), A (Agree), SA (Strongly Agree).}
    \vspace{-0.55cm}
    \label{fig:head_perception_study}
\end{figure*}

\noindent \textbf{Eval. 2: Necessity of Real-Time Mouth Perception -} We perform an ablation study to evaluate the necessity of continuous, real-time mouth perception for the whole duration of inside-mouth bite transfer. This study examines three scenarios, selected based on feedback from individuals with mobility limitations, that impact their feeding process: (1) voluntary head movement, (2) voluntary speaking, and (3) involuntary head movement (spasm).

\textbf{Methods.} In each scenario, we compare two methods: our approach (\textsc{RealTime}), which employs continuous, real-time mouth perception, and the baseline (\textsc{OneTime}) \cite{shaikewitz2023mouth} which estimates the user's mouth pose only once when they first open their mouth. 

\textbf{Procedure.} We conducted the study with 15 participants (11 male, 4 female; ages 20-65) who had no mobility limitations. After participants provide their demographics and complete pre-study questionnaires, they are seated and strapped into a wheelchair, and asked to simulate three scenarios (Figure \ref{fig:head_perception_study}):

S1. Voluntary Head Movement: Participants initially face person A, and open their mouth to initiate a bite. While the robot is moving towards them to feed, upon an audio cue, they turn to person B as if following a conversation. 

S2. Voluntary Speaking: Participants initially face person C, and open their mouth to initiate a bite. While the robot is moving towards them to feed, upon an audio cue, they start reciting a preset script, as if conversing with person C. 

S3. Involuntary Head Movement (Spasm): For this scenario, participants are first trained to mimic involuntary impulses, following guidance from an occupational therapist. During the trial, participants initially face person C, and open their mouth to initiate a bite. While the robot is moving towards them to feed, upon an audio cue, participants simulate a sudden forward impulse. 

Post-feeding, participants fill out a Likert survey on perceived satisfaction, safety, and comfort [all scenarios], control and ease [S1], control and bite timing [S2], and robot responsiveness [S3]. They are fed a total of 24 times (3 scenarios x 2 methods X 2 utensils x 2 trials), with utensil and method orderings counterbalanced.

\textbf{Study Results.} \textsc{RealTime} consistently outperforms \textsc{OneTime} in all scenarios (Figure \ref{fig:head_perception_study}).
In S1, \textsc{OneTime} often misses the target by aiming at the initial mouth position, while \textsc{RealTime} adapts to mouth movements for accurate feeding. In S2, \textsc{OneTime} disrupts speech by continuing movement, whereas \textsc{RealTime} is able to perceive the participant's mouth closing and pauses while they are speaking. In S3, \textsc{OneTime} maintains its movement towards the participant, even if they spasm. In contrast, \textsc{RealTime} immediately retracts upon detecting impulse, resuming feeding only when safe. 

\vspace{-0.25cm}
\subsection{Physical Interaction-Aware Control Evaluation}
\label{study_interaction_classification}
\vspace{-0.1cm}

We compare the performance of various time-series classification models for physical interaction classification, and perform an ablation study to evaluate the necessity of physical interaction-aware control for inside-mouth bite transfer. 


{
\small
\begin{table}[b]
\centering
\vspace{-0.3cm}
\caption{Performance (F1 score) of physical interaction classification methods for inside-mouth bite transfer.}
\vspace{-0.3cm}
\label{tab:contact_model_comparision}
\begin{tabular}{ccccccc}
\hline
\multirow{ 2}{*}{Method} & \multicolumn{3}{c}{All Participants Aggregated} & \multicolumn{3}{c}{Novel Participant} \\
\cmidrule(lr){2-4}        
\cmidrule(lr){5-7}
 & All & Haptic & Visual & All & Haptic & Visual \\ \hline
SVM \cite{cortes1995support} & \textbf{0.903} & \textbf{0.857} & \textbf{0.860} & \textbf{0.872} & 0.834 & 0.649 \\
MLP \cite{haykin1994neural} & 0.892 & 0.845 & 0.842 & 0.871 & \textbf{0.842} & \textbf{0.712} \\
TCN \cite{lea2017temporal} & 0.887 & 0.787 & 0.634 & 0.862 & 0.784 & 0.545 \\ 
TST \cite{zerveas2021tst} & 0.902 & 0.831 & 0.822 & 0.856 & 0.819 & 0.689 \\
\hline 
\vspace{-0.7cm}
\end{tabular}
\end{table}
}

\begin{figure}[b]
    \centering
    \begin{subfigure}[b]{0.47\textwidth}
        \includegraphics[width=\linewidth]{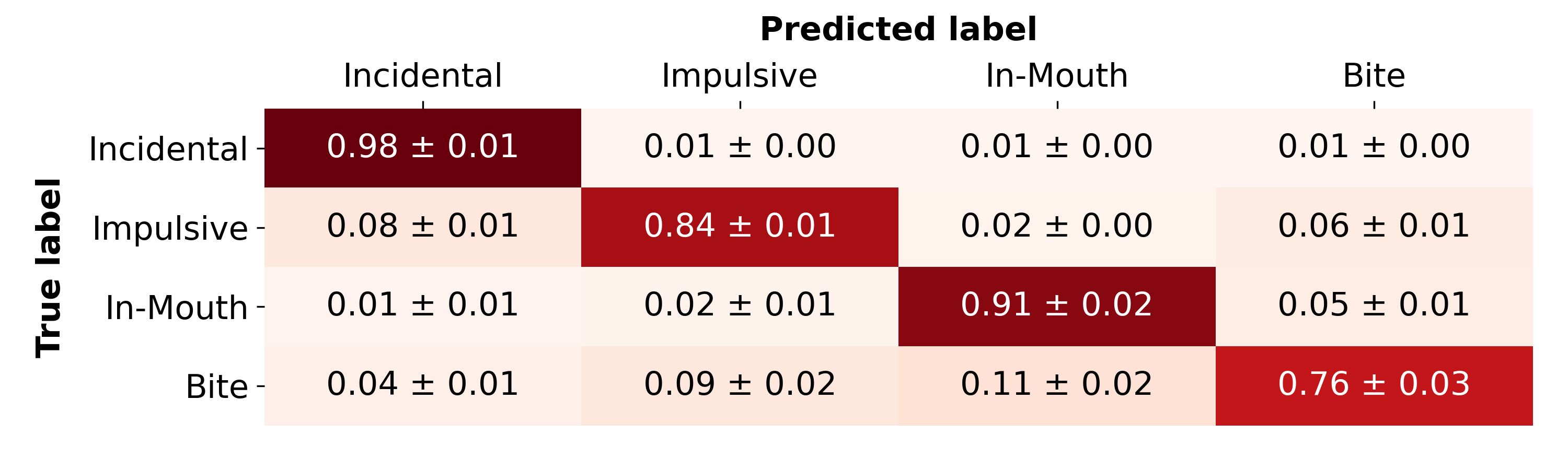}
    \end{subfigure}
    \vspace{-0.55cm}
    \caption{Confusion matrix showing multimodal SVM's performance for Condition B (Novel Participant). }
    \vspace{-0.25cm}
    \label{fig:svm_confusion_matrices}
\end{figure}


\begin{figure*}[t]
    \includegraphics[width=\textwidth]{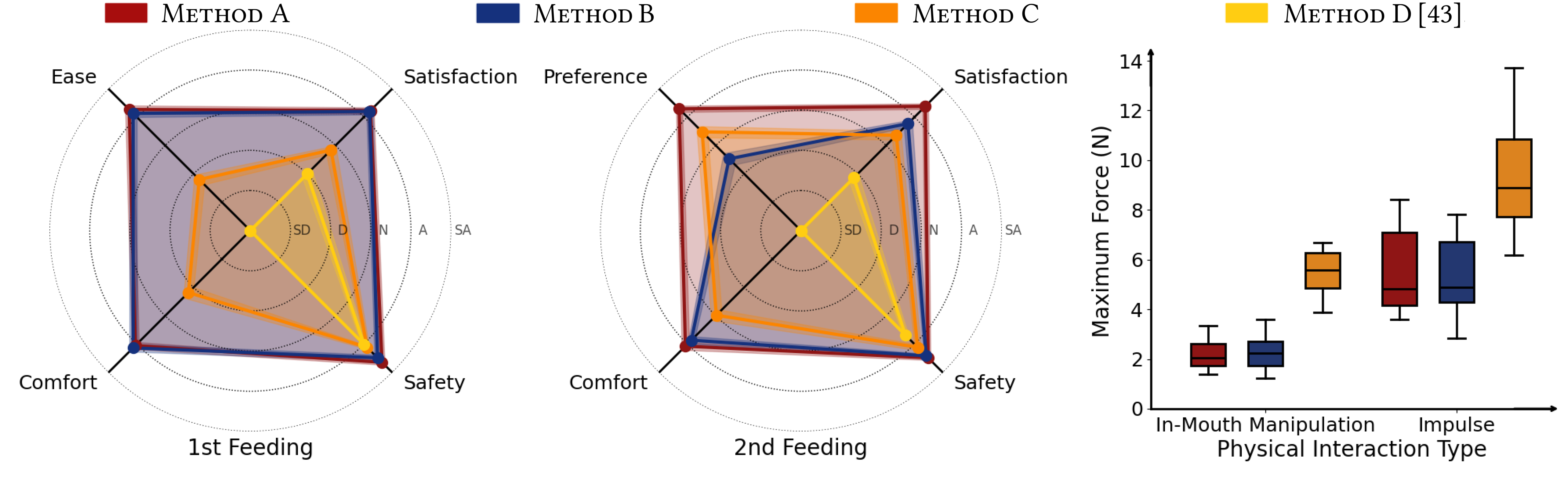}
    \vspace{-0.9cm}
    \caption{More physical-interaction aware control enhances perceived safety, satisfaction and comfort. This is supported by quantitative data recorded by the force sensor. }
    \vspace{-0.6cm}
    \label{fig:contact_classification_study}
\end{figure*}


\noindent \textbf{Eval. 1: Comparison between Models -} We use data collected with the 6 participants in Section \ref{interaction_classification}, and consider two conditions: 
\begin{enumerate}
    \item[A -] Aggregated training and testing datasets from all 6 participants, maintaining an 80:20 split of each participant's data. 
    \item[B -] Train on 5 participants, and test on the 6th, novel participant in leave-one-participant-out cross validation fashion. 
\end{enumerate}
SVM exhibits superior performance across both conditions (Table \ref{tab:contact_model_comparision}), which is likely attributable to the low data regime.
For all models, combining haptic and visual modalities leads to superior performance compared to using either modality in isolation. This finding strongly advocates for the use of multimodal sensing for classifying physical interactions. The reduction in performance from condition A to B could stem from the unique traits of individual participants affecting physical interaction, aligning with findings from previous studies \cite{cioffi2020data, lippi2021data}.

\begin{figure}[b]
    \centering
    \vspace{-0.4cm}
    \includegraphics[width=\linewidth]{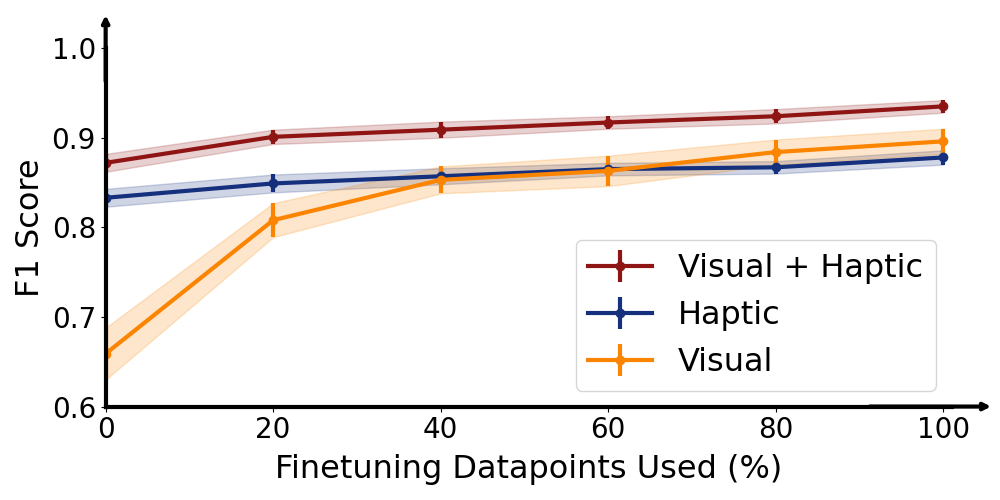} 
    \vspace{-0.9cm}
    \caption{SVM's classification performance on novel participants improves steadily with finetuning using their data.}
    \label{fig:finetune}
    \vspace{-0.2cm}
\end{figure}

\textbf{Performance Improvement with Finetuning.} We investigate finetuning as a solution to the challenge of unique participant traits impacting zero-shot performance of models on novel participants. We split the novel participant's data into an 80:20 ratio, allocating 80\% for finetuning and 20\% for testing. We evaluate the increase in performance of our best performing model, SVM, with incremental amounts of finetuning datapoints. Results (Figure \ref{fig:finetune}) show steady performance improvement across modalities, with a significant performance improvement even with a small number of data points from the novel participant.

\noindent \textbf{Eval. 2: Necessity of Physical Interaction-Aware Control -} We perform an ablation study to assess the importance of physical interaction-aware control for inside-mouth bite transfers, especially for individuals with severe mobility limitations. We consider a scenario where an individual must have food placed on the right side of their mouth due to an inability to bite in the center. When caregivers place food in the center, the person guides them to the correct position using their tongue. Once shown, they expect the caregiver to remember this preference. This person also experiences involuntary head movements, which can occur even while they eat. 

\textbf{Methods.} We manipulate one within-subject variable: the robot's level of physical-interaction awareness during feeding. We use four methods that vary in their ability to distinguish between different contact types, prompting the robot to switch to the appropriate controller accordingly. In order of decreasing awareness:
\begin{itemize}
\item \textsc{Method A}: Incidental vs. Impulsive vs. In-mouth Manipulation vs. Bite classification.
\item \textsc{Method B}: Incidental vs. Inside Mouth Non-bite (combining impulsive and in-mouth manipulation) vs. Bite classification.
\item \textsc{Method C}: Non-bite vs. Bite classification.
\item \textsc{Method D}: Any Contact is Bite \cite{shaikewitz2023mouth}.
\end{itemize}

Methods have access to the ground truth contact type for this study; we also report offline classification accuracy. 

\textbf{Procedure.} 
We conducted the study with 14 participants (8 males, 6 females; ages 21-27) who had no mobility limitations. After providing demographics and completing pre-study questionnaires, participants are seated and strapped into a wheelchair. They then undergo training, under an occupational therapist’s guidance, to realistically simulate the actions of the illustrated care recipient. A trial in this study consists of two successive feedings using the same method, in the following sequence:

1st Feeding: This feeding focuses on in-mouth manipulation. The robot, initially stationary, begins moving when participants open their mouth. Once the robot moves to the center of their mouth, it says "push," prompting participants to move the food to the right side with their tongue while keeping their head still. When the food is correctly positioned, the robot signals "bite," and participants bite down and wait for the robot to withdraw. 

2nd Feeding: This feeding focuses on reaction to impulsive physical interactions and remembering preferred bite transfer location. The robot, initially stationary, begins moving when participants open their mouth. Once inside the mouth, the robot says “impulse,” prompting participants to simulate an impulse. After the movement, they keep their mouth open for five seconds. If the robot doesn't place the food on the right side of their mouth within that time, participants have to adjust it with their tongue. When the food is correctly positioned, the robot signals "bite," and participants bite down and remain still until the robot withdraws. 

Post-feeding, participants complete a Likert survey assessing satisfaction, safety, and comfort for both feedings. Additionally, the survey evaluates ease of moving the utensil for 1st Feeding and whether the robot automatically moved food to the preferred location for 2nd Feeding. Participants are fed a total of 32 times (2 feedings x 4 methods x 2 utensils x 2 trials), with utensil and method orderings counterbalanced.

\textbf{Study Results.} 
Figure \ref{fig:contact_classification_study} presents the study results. 
In the 1st Feeding, which focuses on in-mouth manipulation, participants predominantly prefer \textsc{Method A} and \textsc{Method B}. These methods switch to a force-tracking controller upon detecting in-mouth manipulation, making them easier for participants to manipulate with their tongue compared to \textsc{Method C}, which constantly maintains a goal-tracking controller. This preference is quantitatively supported by the recorded force-torque data. In the 2nd Feeding, centered on reacting to impulsive physical interactions and remembering user preferences, participants rate \textsc{Method A} the highest. Both \textsc{Method A} and \textsc{Method B} transition to a force-tracking controller when impulsive contact is detected, and are perceived as comfortable by participants. However, participants note that \textsc{Method A} remembers their preferred transfer location, unlike \textsc{Method B}, which cannot distinguish between in-mouth manipulation and impulsive contacts. \textsc{Method C}, which continues with the goal-tracking controller during the impulsive motion, is rated as uncomfortable. The quantitative force-torque data show that \textsc{Method C} exerts significantly higher maximum force during an impulse as compared to \textsc{Method A} and \textsc{Method B}. As \textsc{Method C} does not respond to the impulsive contacts, it continues to move towards the preferred position even after the impulse. \textsc{Method D} \cite{shaikewitz2023mouth} retracts from the mouth as soon as any contact is initiated during both 1st Feeding and 2nd Feeding. Consequently, it is ranked as the least satisfying method by the participants.

\textbf{Classification Accuracy on Eval. 2 Study Data.} For this analysis, we use our best-performing model, SVM, trained on 3072 physical interaction data points collected in Section \ref{interaction_classification}, and test on interaction data from this Eval. 2 study.
This evaluation tests the model's generalization capabilities to more unstructured study settings, compared to the controlled conditions of the prior data collection.
SVM achieves an F1 score of 0.719 zero-shot (Table \ref{tab:classification_ablation_study}). 
In addition to distribution shifts caused by the unstructured study setting and novel participants, this performance drop could also be due to changes made to the robot system design between the initial data collection and this ablation study, based on end-user feedback. Inline with improvements seen with finetuning in Eval. 1, the model's performance after further finetuning on 80\% of each participant's data (approximately 10 data points per interaction type) and testing on the remaining 20\% of their data, significantly improves to F1=0.772. 
These results underscore the importance of evaluating finetuning in physical interaction classification models as a means to address various sources of distribution shifts in real-world settings.


All study results are statistically significant (Wilcoxon Signed Rank Test, p<0.05). Details on pilot studies, questions for each measure, and classification results are in the Appendix \cite{appendix}.

\begin{table*}[!b]
\centering
\caption{ Demographics of user study participants. All participants require assistance with feeding.}
\vspace{-0.4cm}
\textbf{BiR - Bite Reflex, GaR - Gag Reflex, LR - Limited head/neck ROM, MaS - Masseter Spasticity, TS - Tongue Spasticity}
\begin{tabular}{lllllll}
\hline
\textbf{ID} & \textbf{Age} & \textbf{Gender} & \textbf{Race} & \textbf{Self-described impairment} & \textbf{Impairment time} & \textbf{Challenges with Feeding} \\ \hline
P1 & 44 & Female & Caucasian/White & Multiple Sclerosis & 25 years & LR, Spasms  \\
P2 & 33 & Female & African American & C3-C4 Spinal Cord Injury & 13 years & LR, Spasms \\
P3 & 30 & Male & Caucasian/White & Arthrogryposis & Since Birth &  GaR, LR, MaS, TS \\
P4 & 45 & Female & Caucasian/White & Multiple Sclerosis & 15 years & LR \\
P5 & 26 & Female & Hispanic & Schizencephaly Quadripelgia & Since Birth & BiR, GaR, LR, MaS, Overbite, TS\\
P6 & 27 & Male & Caucasian/White & Spinal Muscular Atrophy  & Since birth &  LR, MaS \\
P7 & 49 & Female & Hispanic & C4-C5 Spinal Cord Injury & 28 years & LR, Spasms \\
P8 & 44 & Male & Caucasian/White & C4-C5 Spinal Cord Injury & 24 years & LR  \\
P9 & 39 & Female & Asian/Pacific Islander & Spinal Muscular Atrophy & Since Birth & LR, MaS  \\
P10 & 48 & Male & African American & C5-C6 Spinal Cord Injury & 2.5 years & LR, TS  \\
P11 & 31 & Female & African American & Cerebral Palsy Quadriplegia & Since Birth & LR, TS \\
P12 & 25 & Male & Caucasian/White & C4-C5 Spinal Cord Injury & 6 years & LR \\
P13 & 31 & Female & African American & Arthrogryposis & Since Birth & LR, TS \\
\hline
\end{tabular}
\label{Table:study_demographics}
\vspace{-0.2cm} 
\end{table*}
\begin{table}[H]
\centering
\caption{SVM's performance (F1 Score) on Eval. 2 study data.}
\vspace{-0.3cm}
\label{tab:classification_ablation_study}
\begin{tabular}{lccc}
    \hline
     & All & Haptic & Visual \\ \hline
    Zero-Shot & 0.719 & 0.675 & 0.326 \\
    With Finetuning & 0.772 & 0.697 & 0.605 \\
    \hline
\end{tabular}
\vspace{-0.6cm} 
\end{table}




\section{Full System Evaluation with Care Recipients}




We evaluate our full system through a user study with 13 individuals with severe mobility limitations, all of whom require assistance with feeding. The objectives of this study were:

\begin{enumerate}
\item[O1.] To evaluate the effectiveness and acceptance of our inside-mouth bite transfer system among its intended end users.
\item[O2.] Compare participant preferences between inside-mouth and outside-mouth bite transfer \cite{bhattacharjee2020moreautonomy} systems, when both options are functionally possible for an individual.
\end{enumerate}

Due to challenges in recruiting many individuals with mobility limitations at any one place, this study took place at three sites: EmPRISE Lab in Ithaca, NY, Columbia University Medical Center in NYC, NY, and a participant’s home (Figure \ref{fig:user_study}) in Taftville, CT.  
\vspace{-0.2cm}
\subsection{Methods and Procedures}

We considered two systems in this user study: (1) our proposed inside-mouth bite transfer system, 
and (2) an outside-mouth bite transfer system adapted from Bhattacharjee et al. 2020 \cite{bhattacharjee2020moreautonomy} that stops at a fixed distance (5 cms) from the user's mouth.

We informed participants to focus on the interaction from when the robot picks up a food item until they take a bite. Initially, we guided them through practice runs for each method. We used two utensils, a fork for cantaloupe and a spoon for yogurt. A trial involved three successive feedings using the same utensil and bite transfer system, after which we asked questions about their perceived safety and comfort. We evaluated the inside-mouth bite transfer system (O1) with all participants. We compared with the outside-mouth bite transfer system (O2) only with participants who demonstrated the required functional capability during practice runs. Order of the two systems and utensils were counterbalanced. After all trials, a final evaluation phase occurred where we fed participants using inside-mouth bite transfer and asked them to complete a TAM survey \cite{davis1989user}. This was followed by a feedback interview to perceive the significance of the key components of our system and identify areas for improvement.


\subsection{Participants}
\vspace{-0.1cm}

\begin{figure*}[!t]
    \centering
        \includegraphics[width=\linewidth]{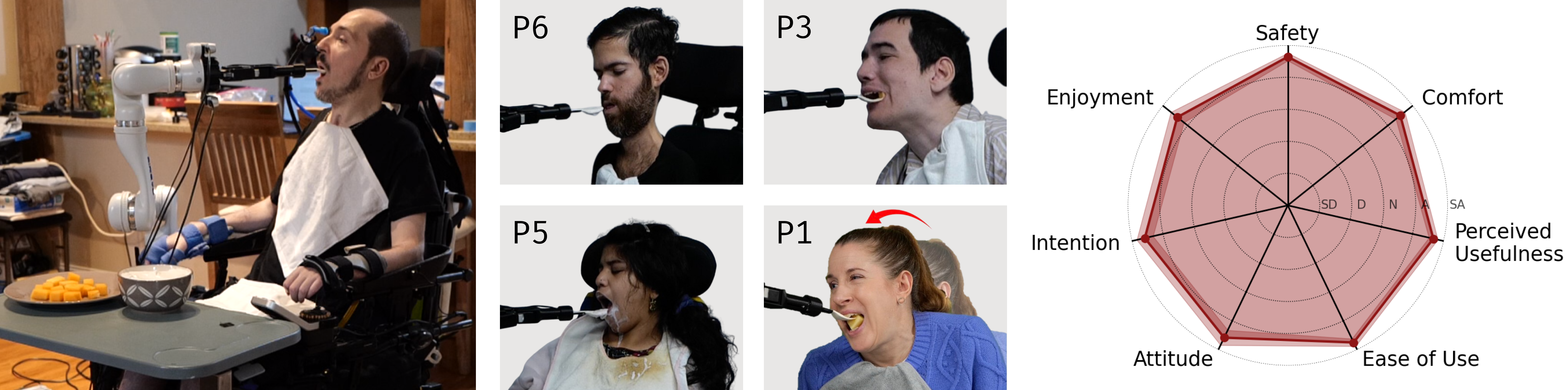}
    \vspace{-0.75cm}
    \caption{Left: Our inside-mouth bite transfer system feeding a care recipient in their home. Center: Inside-mouth bite transfer challenges: P6 has limited mouth opening, P3 can bite only at left molars, P5's weak jaw control causes incidental contacts, and P1's muscle spasms move her towards the utensil. Right: Our system received high ratings from care recipients for its safety and comfort, and was favorably evaluated in terms of technology acceptance, as indicated by the results of the TAM survey. }
    \vspace{-0.5cm}
    \label{fig:user_study}
\end{figure*}

The study involved 13 participants with diverse ages, genders, self-described impairments, duration of impairments, and care providers (Table \ref{Table:study_demographics}). Some participants had complex medical conditions, making feeding particularly challenging (Figure \ref{fig:user_study}). For instance, P6 and P9 have Spinal Muscular Atrophy, a condition known to reduce maximum mouth opening and decrease bite force by almost 50\% \cite{GRANGER1999697}. P3 has an overbite and a sensitive gag reflex, which can be triggered if a utensil presses on their tongue. For solid foods, they prefer the utensil to approach from the left side of their mouth, depositing the food on their left molars. P1, P2 and P7 experience spasms that can happen unexpectedly. 
P5 is diagnosed with Schizencephaly, which manifests with severe motor limitations (quadriplegia) and poor jaw and tongue control for eating. They have severe spasticity in their tongue, gag and bite reflexes, and an overbite.  

All data collection and user studies in this paper were approved by the Cornell University Institutional Review Board.

\vspace{-0.3cm}
\subsection{Findings}
\vspace{-0.05cm}

\hspace{\parindent} \textbf{Our inside-mouth bite transfer system is perceived as safe and comfortable.} Figure \ref{fig:user_study} displays the results. P7 mentioned, "... going in and out, I felt very comfortable, I felt safe, I enjoyed it..." P9 echoed, "... I think I was actually surprised... I liked it, the entrance in and out..." P5's parents were pleasantly surprised, "... I think it worked really much better than I expected. When she moved her mouth, it moved with her... She had fun with it!" P6 emphasized the importance of placement, "The in-the-mouth method worked very well. I was worried before doing it that it would go too deep, and it didn't. I think that making sure it goes in the right place is important, and I think you did very well." 


\textbf{Users view the technology favorably.} As per Figure \ref{fig:user_study}, users found the inside-mouth bite transfer system to be useful, easy to use, have a positive attitude and intend to use it, and enjoy the process. P2's caregiver remarked, "... it will help a lot of people, especially with her condition... it hurts me sometimes when she can't do certain things, so I'm glad." P6's parents highlighted its potential for social engagement, "[in a social setting,] the caregivers all interact, and the kids are just passive recipients of food. So I think this will involve them more in the eating."


\textbf{Even among the few users able to perform outside-mouth bite transfer, some preferred inside-mouth bite transfer.} Out of the 13 users, 9 were unable to lean forward for outside-mouth bite transfer. Among the 4 users who evaluated both systems, some preferred inside-mouth as they felt that outside-mouth bite transfer requires more physical effort. P10 favored inside-mouth, saying it "requires no movement of the neck." P12 and P4's preferences varied based on their physical state. P12 stated, "If I was having neck pain, maybe I would prefer inside mouth so that I am not moving as much." P11 consistently preferred outside-mouth, noting, "I have the ability to position the food in my mouth myself, with my own comfort." 


\textbf{Users highlight that continuous, real-time mouth perception and adaptive compliance are essential.} P12 noted, "I could be sitting in front of the TV watching something... I'm ready to take a bite and then it catches my attention, or yeah, talking to someone." P9 added another context: "When a phone rings, and I turn around..." P7 emphasized the importance of device adaptability during spontaneous movements, "On spasms. I mean, if it wiggles with you, that would be perfect. The body can move when there is a spasm, so it could go off a little and if it follows me, that's great."

\vspace{-0.2cm}
\section{DISCUSSION}

Our study demonstrates promising results, underlining the potential feeding systems have to help improve the quality of life of care recipients. However, the wide range of disabilities and individual user needs makes crafting universal solutions complex. Participants with unique challenges, like limited mouth opening, stressed the importance of personalized adaptations. For instance, P6 and P9, who have smaller mouth openings, suggested specifically designed utensils, with P9 pointing out, "The spoon was too big." Similarly, opinions on transfer speed varied: P11 described the robot as "a little bit too fast," while P12 found it "too slow." These insights highlight the need for adaptive algorithms and interfaces that are tailored to individual needs, moving away from an one-size-fits-all model. Additionally, while our study focuses on short-term interactions, future work needs to explore long-term usability. Longitudinal studies will offer a deeper insight into the prolonged impact of this technology on users.

\vspace{-0.2cm}
\section{ACKNOWLEDGEMENT}

This work was partly funded by NSF IIS \#2132846, CAREER \#2238792, and DARPA under Contract HR001120C0107. We would like to acknowledge Taylor Sanchez, Megan Sofield, Shubhangi Sinha and Skyler Valdez for their help with the user studies.

\bibliographystyle{ACM-Reference-Format}
\bibliography{references/feeding, references/system, references/head_perception, references/physical_interaction, references/medical}


\begin{thebibliography}{48}


\ifx \showCODEN    \undefined \def \showCODEN     #1{\unskip}     \fi
\ifx \showDOI      \undefined \def \showDOI       #1{#1}\fi
\ifx \showISBNx    \undefined \def \showISBNx     #1{\unskip}     \fi
\ifx \showISBNxiii \undefined \def \showISBNxiii  #1{\unskip}     \fi
\ifx \showISSN     \undefined \def \showISSN      #1{\unskip}     \fi
\ifx \showLCCN     \undefined \def \showLCCN      #1{\unskip}     \fi
\ifx \shownote     \undefined \def \shownote      #1{#1}          \fi
\ifx \showarticletitle \undefined \def \showarticletitle #1{#1}   \fi
\ifx \showURL      \undefined \def \showURL       {\relax}        \fi
\providecommand\bibfield[2]{#2}
\providecommand\bibinfo[2]{#2}
\providecommand\natexlab[1]{#1}
\providecommand\showeprint[2][]{arXiv:#2}

\bibitem[ati(2024)]%
        {ati}
 \bibinfo{year}{2024}\natexlab{}.
\newblock \bibinfo{title}{ATI F/T Sensor}.
\newblock
\newblock
\urldef\tempurl%
\url{https://www.ati-ia.com/products/ft/sensors.aspx}
\showURL{%
\tempurl}
\newblock
\shownote{(Accessed: 1st January, 2024)}.


\bibitem[kin(2024)]%
        {kinova}
 \bibinfo{year}{2024}\natexlab{}.
\newblock \bibinfo{title}{Kinova Gen3 6DoF Robotic Arm}.
\newblock
\newblock
\urldef\tempurl%
\url{https://www.kinovarobotics.com/product/gen3-robots}
\showURL{%
\tempurl}
\newblock
\shownote{(Accessed: 1st January, 2024)}.


\bibitem[nea(2024)]%
        {neater}
 \bibinfo{year}{2024}\natexlab{}.
\newblock \bibinfo{title}{Neater Eater Robot}.
\newblock
\newblock
\urldef\tempurl%
\url{https://www.neater.co.uk/neater-eater-robotic}
\showURL{%
\tempurl}
\newblock
\shownote{(Accessed: 1st January, 2024)}.


\bibitem[obi(2024)]%
        {obi}
 \bibinfo{year}{2024}\natexlab{}.
\newblock \bibinfo{title}{Obi}.
\newblock
\newblock
\urldef\tempurl%
\url{https://meetobi.com}
\showURL{%
\tempurl}
\newblock
\shownote{(Accessed: 1st January, 2024)}.


\bibitem[rob(2024)]%
        {robotiq}
 \bibinfo{year}{2024}\natexlab{}.
\newblock \bibinfo{title}{Robotic 2F-85 Gripper}.
\newblock
\newblock
\urldef\tempurl%
\url{https://robotiq.com/products/2f85-140-adaptive-robot-gripper}
\showURL{%
\tempurl}
\newblock
\shownote{(Accessed: 1st January, 2024)}.


\bibitem[app(2024)]%
        {appendix}
 \bibinfo{year}{2024}\natexlab{}.
\newblock \bibinfo{title}{Supplementary Materials}.
\newblock
\newblock
\urldef\tempurl%
\url{https://emprise.cs.cornell.edu/bitetransfer/}
\showURL{%
\tempurl}
\newblock
\shownote{(Accessed: 1st January, 2024)}.


\bibitem[Belkhale et~al\mbox{.}(2021)]%
        {belkhale2021balancing}
\bibfield{author}{\bibinfo{person}{Suneel Belkhale}, \bibinfo{person}{Ethan~K
  Gordon}, \bibinfo{person}{Yuxiao Chen}, \bibinfo{person}{Siddhartha
  Srinivasa}, \bibinfo{person}{Tapomayukh Bhattacharjee}, {and}
  \bibinfo{person}{Dorsa Sadigh}.} \bibinfo{year}{2021}\natexlab{}.
\newblock \showarticletitle{Balancing Efficiency and Comfort in Robot-Assisted
  Bite Transfer}.
\newblock \bibinfo{journal}{\emph{arXiv preprint arXiv:2111.11401}}
  (\bibinfo{year}{2021}).
\newblock


\bibitem[Bhattacharjee et~al\mbox{.}(2020)]%
        {bhattacharjee2020moreautonomy}
\bibfield{author}{\bibinfo{person}{Tapomayukh Bhattacharjee},
  \bibinfo{person}{Ethan~K Gordon}, \bibinfo{person}{Rosario Scalise},
  \bibinfo{person}{Maria~E Cabrera}, \bibinfo{person}{Anat Caspi},
  \bibinfo{person}{Maya Cakmak}, {and} \bibinfo{person}{Siddhartha~S
  Srinivasa}.} \bibinfo{year}{2020}\natexlab{}.
\newblock \showarticletitle{Is more autonomy always better? exploring
  preferences of users with mobility impairments in robot-assisted feeding}. In
  \bibinfo{booktitle}{\emph{2020 15th ACM/IEEE International Conference on
  Human-Robot Interaction (HRI)}}. IEEE, \bibinfo{pages}{181--190}.
\newblock


\bibitem[Booth et~al\mbox{.}(2016)]%
        {booth20163d}
\bibfield{author}{\bibinfo{person}{James Booth}, \bibinfo{person}{Anastasios
  Roussos}, \bibinfo{person}{Stefanos Zafeiriou}, \bibinfo{person}{Allan
  Ponniah}, {and} \bibinfo{person}{David Dunaway}.}
  \bibinfo{year}{2016}\natexlab{}.
\newblock \showarticletitle{A 3d morphable model learnt from 10,000 faces}. In
  \bibinfo{booktitle}{\emph{Proceedings of the IEEE conference on computer
  vision and pattern recognition}}. \bibinfo{pages}{5543--5552}.
\newblock


\bibitem[Brose et~al\mbox{.}(2010)]%
        {brose2010role}
\bibfield{author}{\bibinfo{person}{Steven~W Brose}, \bibinfo{person}{Douglas~J
  Weber}, \bibinfo{person}{Ben~A Salatin}, \bibinfo{person}{Garret~G Grindle},
  \bibinfo{person}{Hongwu Wang}, \bibinfo{person}{Juan~J Vazquez}, {and}
  \bibinfo{person}{Rory~A Cooper}.} \bibinfo{year}{2010}\natexlab{}.
\newblock \showarticletitle{The role of assistive robotics in the lives of
  persons with disability}.
\newblock \bibinfo{journal}{\emph{AJPM\&R}} (\bibinfo{year}{2010}).
\newblock


\bibitem[Candeias et~al\mbox{.}(2018)]%
        {candeias2018vision}
\bibfield{author}{\bibinfo{person}{Alexandre Candeias},
  \bibinfo{person}{Travers Rhodes}, \bibinfo{person}{Manuel Marques},
  \bibinfo{person}{Manuela Veloso}, {et~al\mbox{.}}}
  \bibinfo{year}{2018}\natexlab{}.
\newblock \showarticletitle{Vision augmented robot feeding}. In
  \bibinfo{booktitle}{\emph{Proceedings of the European Conference on Computer
  Vision (ECCV) Workshops}}. \bibinfo{pages}{0--0}.
\newblock


\bibitem[Cao et~al\mbox{.}(2018)]%
        {cao2018vggface2}
\bibfield{author}{\bibinfo{person}{Qiong Cao}, \bibinfo{person}{Li Shen},
  \bibinfo{person}{Weidi Xie}, \bibinfo{person}{Omkar~M Parkhi}, {and}
  \bibinfo{person}{Andrew Zisserman}.} \bibinfo{year}{2018}\natexlab{}.
\newblock \showarticletitle{Vggface2: A dataset for recognising faces across
  pose and age}. In \bibinfo{booktitle}{\emph{2018 13th IEEE international
  conference on automatic face \& gesture recognition (FG 2018)}}. IEEE,
  \bibinfo{pages}{67--74}.
\newblock


\bibitem[Chi{\`o} et~al\mbox{.}(2006)]%
        {chio2006caregiver}
\bibfield{author}{\bibinfo{person}{Adriano Chi{\`o}}, \bibinfo{person}{A
  Gauthier}, \bibinfo{person}{A Vignola}, \bibinfo{person}{Andrea Calvo},
  \bibinfo{person}{Paolo Ghiglione}, \bibinfo{person}{Enrico Cavallo},
  \bibinfo{person}{AA Terreni}, {and} \bibinfo{person}{Roberto Mutani}.}
  \bibinfo{year}{2006}\natexlab{}.
\newblock \showarticletitle{Caregiver time use in ALS}.
\newblock \bibinfo{journal}{\emph{Neurology}} \bibinfo{volume}{67},
  \bibinfo{number}{5} (\bibinfo{year}{2006}), \bibinfo{pages}{902--904}.
\newblock


\bibitem[Chung et~al\mbox{.}(2018)]%
        {chung2018voxceleb2}
\bibfield{author}{\bibinfo{person}{J Chung}, \bibinfo{person}{A Nagrani}, {and}
  \bibinfo{person}{A Zisserman}.} \bibinfo{year}{2018}\natexlab{}.
\newblock \showarticletitle{VoxCeleb2: Deep speaker recognition}.
\newblock \bibinfo{journal}{\emph{Interspeech 2018}} (\bibinfo{year}{2018}).
\newblock


\bibitem[Cioffi et~al\mbox{.}(2020)]%
        {cioffi2020data}
\bibfield{author}{\bibinfo{person}{Giovanni Cioffi}, \bibinfo{person}{Silke
  Klose}, {and} \bibinfo{person}{Arne Wahrburg}.}
  \bibinfo{year}{2020}\natexlab{}.
\newblock \showarticletitle{Data-efficient online classification of human-robot
  contact situations}. In \bibinfo{booktitle}{\emph{2020 European Control
  Conference (ECC)}}. IEEE, \bibinfo{pages}{608--614}.
\newblock


\bibitem[Cortes and Vapnik(1995)]%
        {cortes1995support}
\bibfield{author}{\bibinfo{person}{Corinna Cortes} {and}
  \bibinfo{person}{Vladimir Vapnik}.} \bibinfo{year}{1995}\natexlab{}.
\newblock \showarticletitle{Support-vector networks}.
\newblock \bibinfo{journal}{\emph{Machine learning}} \bibinfo{volume}{20},
  \bibinfo{number}{3} (\bibinfo{year}{1995}), \bibinfo{pages}{273--297}.
\newblock


\bibitem[Dai et~al\mbox{.}(2017)]%
        {dai20173d}
\bibfield{author}{\bibinfo{person}{Hang Dai}, \bibinfo{person}{Nick Pears},
  \bibinfo{person}{William~AP Smith}, {and} \bibinfo{person}{Christian
  Duncan}.} \bibinfo{year}{2017}\natexlab{}.
\newblock \showarticletitle{A 3d morphable model of craniofacial shape and
  texture variation}. In \bibinfo{booktitle}{\emph{Proceedings of the IEEE
  international conference on computer vision}}. \bibinfo{pages}{3085--3093}.
\newblock


\bibitem[Davis et~al\mbox{.}(1989)]%
        {davis1989user}
\bibfield{author}{\bibinfo{person}{Fred~D Davis}, \bibinfo{person}{Richard~P
  Bagozzi}, {and} \bibinfo{person}{Paul~R Warshaw}.}
  \bibinfo{year}{1989}\natexlab{}.
\newblock \showarticletitle{User acceptance of computer technology: A
  comparison of two theoretical models}.
\newblock \bibinfo{journal}{\emph{Management science}} \bibinfo{volume}{35},
  \bibinfo{number}{8} (\bibinfo{year}{1989}), \bibinfo{pages}{982--1003}.
\newblock


\bibitem[Dreer et~al\mbox{.}(2007)]%
        {dreer2007family}
\bibfield{author}{\bibinfo{person}{Laura~E Dreer}, \bibinfo{person}{Timothy~R
  Elliott}, \bibinfo{person}{Richard Shewchuk}, \bibinfo{person}{Jack~W Berry},
  {and} \bibinfo{person}{Patricia Rivera}.} \bibinfo{year}{2007}\natexlab{}.
\newblock \showarticletitle{Family caregivers of persons with spinal cord
  injury: Predicting caregivers at risk for probable depression.}
\newblock \bibinfo{journal}{\emph{Rehabilitation Psychology}}
  \bibinfo{volume}{52}, \bibinfo{number}{3} (\bibinfo{year}{2007}),
  \bibinfo{pages}{351}.
\newblock


\bibitem[Fazeli et~al\mbox{.}(2019)]%
        {fazeli2019see}
\bibfield{author}{\bibinfo{person}{Nima Fazeli}, \bibinfo{person}{Miquel
  Oller}, \bibinfo{person}{Jiajun Wu}, \bibinfo{person}{Zheng Wu},
  \bibinfo{person}{Joshua~B Tenenbaum}, {and} \bibinfo{person}{Alberto
  Rodriguez}.} \bibinfo{year}{2019}\natexlab{}.
\newblock \showarticletitle{See, feel, act: Hierarchical learning for complex
  manipulation skills with multisensory fusion}.
\newblock \bibinfo{journal}{\emph{Science Robotics}} \bibinfo{volume}{4},
  \bibinfo{number}{26} (\bibinfo{year}{2019}), \bibinfo{pages}{eaav3123}.
\newblock


\bibitem[Feng et~al\mbox{.}(2019)]%
        {feng2019robot}
\bibfield{author}{\bibinfo{person}{Ryan Feng}, \bibinfo{person}{Youngsun Kim},
  \bibinfo{person}{Gilwoo Lee}, \bibinfo{person}{Ethan~K Gordon},
  \bibinfo{person}{Matt Schmittle}, \bibinfo{person}{Shivaum Kumar},
  \bibinfo{person}{Tapomayukh Bhattacharjee}, {and}
  \bibinfo{person}{Siddhartha~S Srinivasa}.} \bibinfo{year}{2019}\natexlab{}.
\newblock \showarticletitle{Robot-assisted feeding: Generalizing skewering
  strategies across food items on a realistic plate}.
\newblock \bibinfo{journal}{\emph{arXiv preprint arXiv:1906.02350}}
  (\bibinfo{year}{2019}).
\newblock


\bibitem[Feng et~al\mbox{.}(2021)]%
        {feng2021learning}
\bibfield{author}{\bibinfo{person}{Yao Feng}, \bibinfo{person}{Haiwen Feng},
  \bibinfo{person}{Michael~J Black}, {and} \bibinfo{person}{Timo Bolkart}.}
  \bibinfo{year}{2021}\natexlab{}.
\newblock \showarticletitle{Learning an animatable detailed 3D face model from
  in-the-wild images}.
\newblock \bibinfo{journal}{\emph{ACM Transactions on Graphics (TOG)}}
  (\bibinfo{year}{2021}), \bibinfo{pages}{1--13}.
\newblock


\bibitem[Franzel et~al\mbox{.}(2021)]%
        {franzel2021detection}
\bibfield{author}{\bibinfo{person}{Felix Franzel}, \bibinfo{person}{Thomas
  Eiband}, {and} \bibinfo{person}{Dongheui Lee}.}
  \bibinfo{year}{2021}\natexlab{}.
\newblock \showarticletitle{Detection of Collaboration and Collision Events
  during Contact Task Execution}. In \bibinfo{booktitle}{\emph{2020 IEEE-RAS
  20th International Conference on Humanoid Robots (Humanoids)}}. IEEE,
  \bibinfo{pages}{376--383}.
\newblock


\bibitem[Gallenberger et~al\mbox{.}(2019)]%
        {gallenberger2019transfer}
\bibfield{author}{\bibinfo{person}{Daniel Gallenberger},
  \bibinfo{person}{Tapomayukh Bhattacharjee}, \bibinfo{person}{Youngsun Kim},
  {and} \bibinfo{person}{Siddhartha~S Srinivasa}.}
  \bibinfo{year}{2019}\natexlab{}.
\newblock \showarticletitle{Transfer depends on acquisition: Analyzing
  manipulation strategies for robotic feeding}. In
  \bibinfo{booktitle}{\emph{2019 14th ACM/IEEE International Conference on
  Human-Robot Interaction (HRI)}}. IEEE, \bibinfo{pages}{267--276}.
\newblock


\bibitem[Golz et~al\mbox{.}(2015)]%
        {golz2015using}
\bibfield{author}{\bibinfo{person}{Saskia Golz}, \bibinfo{person}{Christian
  Osendorfer}, {and} \bibinfo{person}{Sami Haddadin}.}
  \bibinfo{year}{2015}\natexlab{}.
\newblock \showarticletitle{Using tactile sensation for learning contact
  knowledge: Discriminate collision from physical interaction}. In
  \bibinfo{booktitle}{\emph{2015 IEEE International Conference on Robotics and
  Automation (ICRA)}}. IEEE, \bibinfo{pages}{3788--3794}.
\newblock


\bibitem[Gordon et~al\mbox{.}(2020)]%
        {gordon2020adaptive}
\bibfield{author}{\bibinfo{person}{Ethan~K Gordon}, \bibinfo{person}{Xiang
  Meng}, \bibinfo{person}{Tapomayukh Bhattacharjee}, \bibinfo{person}{Matt
  Barnes}, {and} \bibinfo{person}{Siddhartha~S Srinivasa}.}
  \bibinfo{year}{2020}\natexlab{}.
\newblock \showarticletitle{Adaptive robot-assisted feeding: An online learning
  framework for acquiring previously unseen food items}. In
  \bibinfo{booktitle}{\emph{2020 IEEE/RSJ International Conference on
  Intelligent Robots and Systems (IROS)}}. IEEE, \bibinfo{pages}{9659--9666}.
\newblock


\bibitem[Gordon et~al\mbox{.}(2023)]%
        {gordon2023towards}
\bibfield{author}{\bibinfo{person}{Ethan~Kroll Gordon}, \bibinfo{person}{Amal
  Nanavati}, \bibinfo{person}{Ramya Challa}, \bibinfo{person}{Bernie~Hao Zhu},
  \bibinfo{person}{Taylor Annette~Kessler Faulkner}, {and}
  \bibinfo{person}{Siddhartha Srinivasa}.} \bibinfo{year}{2023}\natexlab{}.
\newblock \showarticletitle{Towards General Single-Utensil Food Acquisition
  with Human-Informed Actions}. In \bibinfo{booktitle}{\emph{Conference on
  Robot Learning}}. PMLR, \bibinfo{pages}{2414--2428}.
\newblock


\bibitem[Gordon et~al\mbox{.}(2021)]%
        {gordon2021leveraging}
\bibfield{author}{\bibinfo{person}{Ethan~K Gordon}, \bibinfo{person}{Sumegh
  Roychowdhury}, \bibinfo{person}{Tapomayukh Bhattacharjee},
  \bibinfo{person}{Kevin Jamieson}, {and} \bibinfo{person}{Siddhartha~S
  Srinivasa}.} \bibinfo{year}{2021}\natexlab{}.
\newblock \showarticletitle{Leveraging Post Hoc Context for Faster Learning in
  Bandit Settings with Applications in Robot-Assisted Feeding}. In
  \bibinfo{booktitle}{\emph{2021 IEEE International Conference on Robotics and
  Automation (ICRA)}}. IEEE, \bibinfo{pages}{10528--10535}.
\newblock


\bibitem[Granger et~al\mbox{.}(1999)]%
        {GRANGER1999697}
\bibfield{author}{\bibinfo{person}{M.W. Granger}, \bibinfo{person}{P.H.
  Buschang}, \bibinfo{person}{G.S. Throckmorton}, {and} \bibinfo{person}{S.T.
  Iannaccone}.} \bibinfo{year}{1999}\natexlab{}.
\newblock \showarticletitle{Masticatory muscle function in patients with spinal
  muscular atrophy}.
\newblock \bibinfo{journal}{\emph{American Journal of Orthodontics and
  Dentofacial Orthopedics}} \bibinfo{volume}{115}, \bibinfo{number}{6}
  (\bibinfo{year}{1999}), \bibinfo{pages}{697--702}.
\newblock
\showISSN{0889-5406}
\urldef\tempurl%
\url{https://doi.org/10.1016/S0889-5406(99)70296-9}
\showDOI{\tempurl}


\bibitem[Haykin(1994)]%
        {haykin1994neural}
\bibfield{author}{\bibinfo{person}{Simon Haykin}.}
  \bibinfo{year}{1994}\natexlab{}.
\newblock \bibinfo{booktitle}{\emph{Neural networks: a comprehensive
  foundation}}.
\newblock \bibinfo{publisher}{Prentice Hall PTR}.
\newblock


\bibitem[Jacobsson et~al\mbox{.}(2000)]%
        {jacobsson2000eatingprocess}
\bibfield{author}{\bibinfo{person}{Catrine Jacobsson}, \bibinfo{person}{Karin
  Axelsson}, \bibinfo{person}{Per~Olov Österlind}, {and}
  \bibinfo{person}{Astrid Norberg}.} \bibinfo{year}{2000}\natexlab{}.
\newblock \showarticletitle{How people with stroke and healthy older people
  experience the eating process}.
\newblock \bibinfo{journal}{\emph{Journal of Clinical Nursing}}
  \bibinfo{volume}{9}, \bibinfo{number}{2} (\bibinfo{year}{2000}),
  \bibinfo{pages}{255--264}.
\newblock
\urldef\tempurl%
\url{https://doi.org/10.1046/j.1365-2702.2000.00355.x}
\showDOI{\tempurl}


\bibitem[Katz et~al\mbox{.}(1963)]%
        {katz1963studies}
\bibfield{author}{\bibinfo{person}{Sidney Katz}, \bibinfo{person}{Amasa~B
  Ford}, \bibinfo{person}{Roland~W Moskowitz}, \bibinfo{person}{Beverly~A
  Jackson}, {and} \bibinfo{person}{Marjorie~W Jaffe}.}
  \bibinfo{year}{1963}\natexlab{}.
\newblock \showarticletitle{Studies of illness in the aged: the index of ADL: a
  standardized measure of biological and psychosocial function}.
\newblock \bibinfo{journal}{\emph{jama}} \bibinfo{volume}{185},
  \bibinfo{number}{12} (\bibinfo{year}{1963}), \bibinfo{pages}{914--919}.
\newblock


\bibitem[Kouris et~al\mbox{.}(2018)]%
        {kouris2018frequency}
\bibfield{author}{\bibinfo{person}{Alexandros Kouris}, \bibinfo{person}{Fotios
  Dimeas}, {and} \bibinfo{person}{Nikos Aspragathos}.}
  \bibinfo{year}{2018}\natexlab{}.
\newblock \showarticletitle{A frequency domain approach for contact type
  distinction in human--robot collaboration}.
\newblock \bibinfo{journal}{\emph{IEEE robotics and automation letters}}
  \bibinfo{volume}{3}, \bibinfo{number}{2} (\bibinfo{year}{2018}),
  \bibinfo{pages}{720--727}.
\newblock


\bibitem[Lea et~al\mbox{.}(2017)]%
        {lea2017temporal}
\bibfield{author}{\bibinfo{person}{Colin Lea}, \bibinfo{person}{Michael~D
  Flynn}, \bibinfo{person}{Rene Vidal}, \bibinfo{person}{Austin Reiter}, {and}
  \bibinfo{person}{Gregory~D Hager}.} \bibinfo{year}{2017}\natexlab{}.
\newblock \showarticletitle{Temporal convolutional networks for action
  segmentation and detection}. In \bibinfo{booktitle}{\emph{proceedings of the
  IEEE Conference on Computer Vision and Pattern Recognition}}.
  \bibinfo{pages}{156--165}.
\newblock


\bibitem[Lee et~al\mbox{.}(2019)]%
        {lee2019making}
\bibfield{author}{\bibinfo{person}{Michelle~A Lee}, \bibinfo{person}{Yuke Zhu},
  \bibinfo{person}{Krishnan Srinivasan}, \bibinfo{person}{Parth Shah},
  \bibinfo{person}{Silvio Savarese}, \bibinfo{person}{Li Fei-Fei},
  \bibinfo{person}{Animesh Garg}, {and} \bibinfo{person}{Jeannette Bohg}.}
  \bibinfo{year}{2019}\natexlab{}.
\newblock \showarticletitle{Making sense of vision and touch: Self-supervised
  learning of multimodal representations for contact-rich tasks}. In
  \bibinfo{booktitle}{\emph{2019 International Conference on Robotics and
  Automation (ICRA)}}. IEEE, \bibinfo{pages}{8943--8950}.
\newblock


\bibitem[Li et~al\mbox{.}(2017)]%
        {li2017flame}
\bibfield{author}{\bibinfo{person}{Tianye Li}, \bibinfo{person}{Timo Bolkart},
  \bibinfo{person}{Michael~J Black}, \bibinfo{person}{Hao Li}, {and}
  \bibinfo{person}{Javier Romero}.} \bibinfo{year}{2017}\natexlab{}.
\newblock \showarticletitle{Learning a model of facial shape and expression
  from 4D scans.}
\newblock \bibinfo{journal}{\emph{ACM Trans. Graph.}} \bibinfo{volume}{36},
  \bibinfo{number}{6} (\bibinfo{year}{2017}), \bibinfo{pages}{194--1}.
\newblock


\bibitem[Li et~al\mbox{.}(2019)]%
        {li2019connecting}
\bibfield{author}{\bibinfo{person}{Yunzhu Li}, \bibinfo{person}{Jun-Yan Zhu},
  \bibinfo{person}{Russ Tedrake}, {and} \bibinfo{person}{Antonio Torralba}.}
  \bibinfo{year}{2019}\natexlab{}.
\newblock \showarticletitle{Connecting touch and vision via cross-modal
  prediction}. In \bibinfo{booktitle}{\emph{Proceedings of the IEEE/CVF
  Conference on Computer Vision and Pattern Recognition}}.
  \bibinfo{pages}{10609--10618}.
\newblock


\bibitem[Lippi et~al\mbox{.}(2021)]%
        {lippi2021data}
\bibfield{author}{\bibinfo{person}{Martina Lippi}, \bibinfo{person}{Giuseppe
  Gillini}, \bibinfo{person}{Alessandro Marino}, {and} \bibinfo{person}{Filippo
  Arrichiello}.} \bibinfo{year}{2021}\natexlab{}.
\newblock \showarticletitle{A Data-Driven Approach for Contact Detection,
  Classification and Reaction in Physical Human-Robot Collaboration}. In
  \bibinfo{booktitle}{\emph{2021 IEEE International Conference on Robotics and
  Automation (ICRA)}}. IEEE, \bibinfo{pages}{3597--3603}.
\newblock


\bibitem[Lynch and Cahalan(2017)]%
        {lynch2017impact}
\bibfield{author}{\bibinfo{person}{J Lynch} {and} \bibinfo{person}{R Cahalan}.}
  \bibinfo{year}{2017}\natexlab{}.
\newblock \showarticletitle{The impact of spinal cord injury on the quality of
  life of primary family caregivers: a literature review}.
\newblock \bibinfo{journal}{\emph{Spinal cord}} \bibinfo{volume}{55},
  \bibinfo{number}{11} (\bibinfo{year}{2017}), \bibinfo{pages}{964--978}.
\newblock


\bibitem[Madan et~al\mbox{.}(2022)]%
        {madan2022sparcs}
\bibfield{author}{\bibinfo{person}{Rishabh Madan}, \bibinfo{person}{Rajat~Kumar
  Jenamani}, \bibinfo{person}{Vy~Thuy Nguyen}, \bibinfo{person}{Ahmed
  Moustafa}, \bibinfo{person}{Xuefeng Hu}, \bibinfo{person}{Katherine
  Dimitropoulou}, {and} \bibinfo{person}{Tapomayukh Bhattacharjee}.}
  \bibinfo{year}{2022}\natexlab{}.
\newblock \showarticletitle{Sparcs: Structuring physically assistive robotics
  for caregiving with stakeholders-in-the-loop}. In
  \bibinfo{booktitle}{\emph{2022 IEEE/RSJ International Conference on
  Intelligent Robots and Systems (IROS)}}. IEEE, \bibinfo{pages}{641--648}.
\newblock


\bibitem[Nanavati et~al\mbox{.}(2023)]%
        {nanavati2023design}
\bibfield{author}{\bibinfo{person}{Amal Nanavati}, \bibinfo{person}{Patricia
  Alves-Oliveira}, \bibinfo{person}{Tyler Schrenk}, \bibinfo{person}{Ethan~K
  Gordon}, \bibinfo{person}{Maya Cakmak}, {and} \bibinfo{person}{Siddhartha~S
  Srinivasa}.} \bibinfo{year}{2023}\natexlab{}.
\newblock \showarticletitle{Design principles for robot-assisted feeding in
  social contexts}. In \bibinfo{booktitle}{\emph{Proceedings of the 2023
  ACM/IEEE International Conference on Human-Robot Interaction}}.
  \bibinfo{pages}{24--33}.
\newblock


\bibitem[Park et~al\mbox{.}(2020)]%
        {park2020evaluation}
\bibfield{author}{\bibinfo{person}{Daehyung Park}, \bibinfo{person}{Yuuna
  Hoshi}, \bibinfo{person}{Harshal~P. Mahajan}, \bibinfo{person}{Ho~Keun Kim},
  \bibinfo{person}{Zackory Erickson}, \bibinfo{person}{Wendy~A. Rogers}, {and}
  \bibinfo{person}{Charles~C. Kemp}.} \bibinfo{year}{2020}\natexlab{}.
\newblock \showarticletitle{Active robot-assisted feeding with a
  general-purpose mobile manipulator: Design, evaluation, and lessons learned}.
\newblock \bibinfo{journal}{\emph{Robotics and Autonomous Systems}}
  \bibinfo{volume}{124} (\bibinfo{year}{2020}), \bibinfo{pages}{103344}.
\newblock
\showISSN{0921-8890}
\urldef\tempurl%
\url{https://doi.org/10.1016/j.robot.2019.103344}
\showDOI{\tempurl}


\bibitem[Shaikewitz et~al\mbox{.}(2023)]%
        {shaikewitz2023mouth}
\bibfield{author}{\bibinfo{person}{Lorenzo Shaikewitz}, \bibinfo{person}{Yilin
  Wu}, \bibinfo{person}{Suneel Belkhale}, \bibinfo{person}{Jennifer Grannen},
  \bibinfo{person}{Priya Sundaresan}, {and} \bibinfo{person}{Dorsa Sadigh}.}
  \bibinfo{year}{2023}\natexlab{}.
\newblock \showarticletitle{In-Mouth Robotic Bite Transfer with Visual and
  Haptic Sensing}. In \bibinfo{booktitle}{\emph{2023 IEEE International
  Conference on Robotics and Automation (ICRA)}}. IEEE,
  \bibinfo{pages}{9885--9895}.
\newblock


\bibitem[Shune(2020)]%
        {shune2020experience}
\bibfield{author}{\bibinfo{person}{Samantha~E. Shune}.}
  \bibinfo{year}{2020}\natexlab{}.
\newblock \showarticletitle{An Altered Eating Experience: Attitudes Toward
  Feeding Assistance Among Younger and Older Adults}.
\newblock \bibinfo{journal}{\emph{Rehabilitation nursing : the official journal
  of the Association of Rehabilitation Nurses}} (\bibinfo{year}{2020}).
\newblock


\bibitem[Sundaresan et~al\mbox{.}(2023)]%
        {sundaresan2023learning}
\bibfield{author}{\bibinfo{person}{Priya Sundaresan}, \bibinfo{person}{Suneel
  Belkhale}, {and} \bibinfo{person}{Dorsa Sadigh}.}
  \bibinfo{year}{2023}\natexlab{}.
\newblock \showarticletitle{Learning visuo-haptic skewering strategies for
  robot-assisted feeding}. In \bibinfo{booktitle}{\emph{Conference on Robot
  Learning}}. PMLR, \bibinfo{pages}{332--341}.
\newblock


\bibitem[Taylor(2018)]%
        {taylor2018americans}
\bibfield{author}{\bibinfo{person}{Danielle~M Taylor}.}
  \bibinfo{year}{2018}\natexlab{}.
\newblock \showarticletitle{Americans with disabilities: 2014}.
\newblock \bibinfo{journal}{\emph{US Census Bureau}} (\bibinfo{year}{2018}),
  \bibinfo{pages}{1--32}.
\newblock


\bibitem[Wang et~al\mbox{.}(2019)]%
        {wang2019racial}
\bibfield{author}{\bibinfo{person}{Mei Wang}, \bibinfo{person}{Weihong Deng},
  \bibinfo{person}{Jiani Hu}, \bibinfo{person}{Xunqiang Tao}, {and}
  \bibinfo{person}{Yaohai Huang}.} \bibinfo{year}{2019}\natexlab{}.
\newblock \showarticletitle{Racial faces in the wild: Reducing racial bias by
  information maximization adaptation network}. In
  \bibinfo{booktitle}{\emph{Proceedings of the ieee/cvf international
  conference on computer vision}}. \bibinfo{pages}{692--702}.
\newblock


\bibitem[Zerveas et~al\mbox{.}(2021)]%
        {zerveas2021tst}
\bibfield{author}{\bibinfo{person}{George Zerveas}, \bibinfo{person}{Srideepika
  Jayaraman}, \bibinfo{person}{Dhaval Patel}, \bibinfo{person}{Anuradha
  Bhamidipaty}, {and} \bibinfo{person}{Carsten Eickhoff}.}
  \bibinfo{year}{2021}\natexlab{}.
\newblock \showarticletitle{A Transformer-Based Framework for Multivariate Time
  Series Representation Learning}. In \bibinfo{booktitle}{\emph{Proceedings of
  the 27th ACM SIGKDD Conference on Knowledge Discovery \&amp; Data Mining}}
  (Virtual Event, Singapore) \emph{(\bibinfo{series}{KDD '21})}.
  \bibinfo{publisher}{Association for Computing Machinery},
  \bibinfo{address}{New York, NY, USA}, \bibinfo{pages}{2114–2124}.
\newblock
\showISBNx{9781450383325}
\urldef\tempurl%
\url{https://doi.org/10.1145/3447548.3467401}
\showDOI{\tempurl}


\end{thebibliography}

\end{document}